\documentclass[journal]{IEEEtran}

\usepackage{cite}
\usepackage{amsmath,amssymb,amsfonts}
\usepackage{algorithmic}
\usepackage{mathrsfs}
\usepackage{graphicx}
\usepackage{textcomp}
\usepackage{xcolor}
\usepackage{booktabs}
\usepackage{multirow}
\usepackage{colortbl}
\usepackage[colorlinks,
            linkcolor=red,
            anchorcolor=blue,
            citecolor=green
            ]{hyperref}

\begin{document}
%
\title{Lightweight Real-time Semantic Segmentation Network with Efficient Transformer and CNN}
%
%
%
\author{Guoan Xu,
        Juncheng Li,
        Guangwei Gao,~\IEEEmembership{Senior Member,~IEEE,}
        Huimin Lu,~\IEEEmembership{Senior Member,~IEEE}\\
        Jian Yang,~\IEEEmembership{Member,~IEEE}
        and Dong Yue,~\IEEEmembership{Fellow,~IEEE,}
\thanks{This work was partly supported by the National Natural Science Foundation of China under Grants 61972212 and 61833011, and the Open Fund Project of Provincial Key Laboratory for Computer Information Processing Technology (Soochow University) under Grant KJS2274. \textit{(Guoan Xu, Juncheng Li, and Guangwei Gao contributed equally to this work.) (Corresponding author: Guangwei Gao.)}}
\thanks{Guoan Xu, Guangwei Gao, and Dong Yue are with the Institute of Advanced Technology, Nanjing University of Posts and Telecommunications, Nanjing 210023, China, and also with the Provincial Key Laboratory for Computer Information Processing Technology, Soochow University, Suzhou 215006, China (e-mail: xga\_njupt@163.com, csggao@gmail.com, medongy@vip.163.com).}
\thanks{Juncheng Li is with the School of Communication and Information Engineering, Shanghai University, Shanghai 200444, China, also with Jiangsu Key Laboratory of Image and Video Understanding for Social Safety, Nanjing University of Science and Technology, Nanjing 210049, China (e-mail: cvjunchengli@gmail.com).}
\thanks{Huimin Lu is with the Kyushu Institute of Technology, Kitakyushu 804-8550, Japan (e-mail: dr.huimin.lu@ieee.org).}
\thanks{Jian Yang is with the School of Computer Science and Technology, Nanjing University of Science and Technology, Nanjing 210049, China (e-mail: csjyang@njust.edu.cn).}
}

\markboth{IEEE Transactions on Intelligent Transportation Systems}%
{Shell \MakeLowercase{\textit{et al.}}: Bare Demo of IEEEtran.cls for IEEE Journals}
%

\maketitle

\begin{abstract}
In the past decade, convolutional neural networks (CNNs) have shown prominence for semantic segmentation. Although CNN models have very impressive performance, the ability to capture global representation is still insufficient, which results in suboptimal results. Recently, Transformer achieved huge success in NLP tasks, demonstrating its advantages in modeling long-range dependency. Recently, Transformer has also attracted tremendous attention from computer vision researchers who reformulate the image processing tasks as a sequence-to-sequence prediction but resulted in deteriorating local feature details. In this work, we propose a lightweight real-time semantic segmentation network called LETNet. LETNet combines a U-shaped CNN with Transformer effectively in a capsule embedding style to compensate for respective deficiencies. Meanwhile, the elaborately designed Lightweight Dilated Bottleneck (LDB) module and Feature Enhancement (FE) module cultivate a positive impact on training from scratch simultaneously. Extensive experiments performed on challenging datasets demonstrate that LETNet achieves superior performances in accuracy and efficiency balance. Specifically, It only contains 0.95M parameters and 13.6G FLOPs but yields 72.8\% mIoU at 120 FPS on the Cityscapes test set and 70.5\% mIoU at 250 FPS on the CamVid test dataset using a single RTX 3090 GPU. Source code will be available at \textit{\url{https://github.com/IVIPLab/LETNet}}.
\end{abstract}


%
\IEEEpeerreviewmaketitle

\begin{figure}[htbp]
	\centerline{\includegraphics[width=9.5cm]{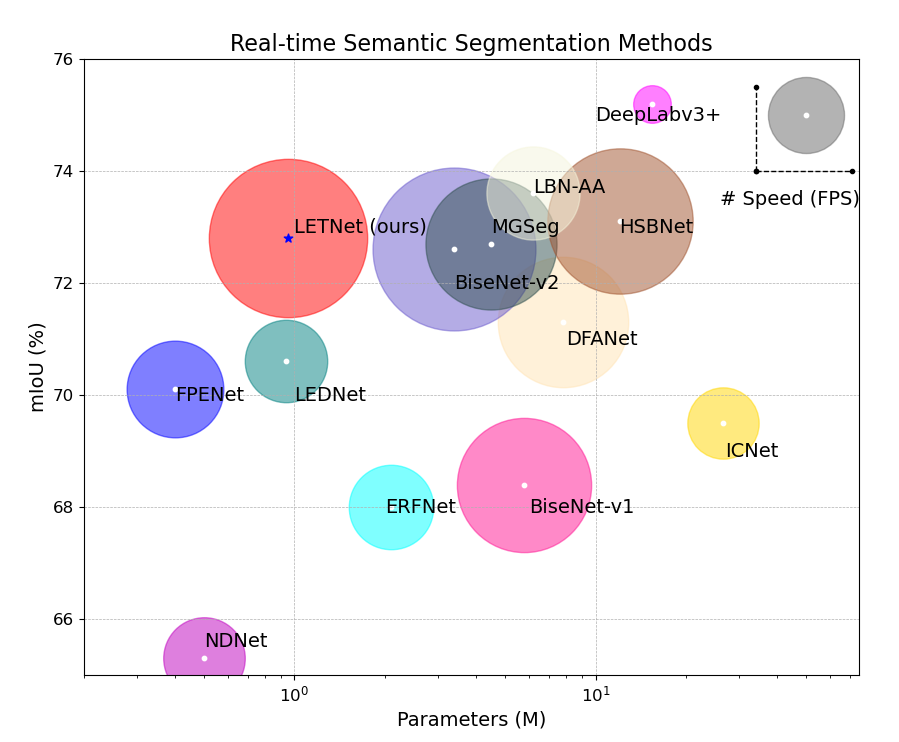}}
	\caption{Accuracy-Parameters-Speed evaluations on the Cityscapes test dataset. A larger radius of a circle indicates a faster inference speed.}
	\label{Figure1}
\end{figure}

\section{Introduction}
The task of semantic segmentation aims to assign a semantic label to each pixel, which is widely used in augmented reality devices, autonomous driving, and video surveillance. Since the Fully Convolutional Network (FCN~\cite{long2015fully}) was proposed, existing semantic segmentation models have been using it as a prototype for improvement. However, the receptive field of FCN-based models is limited. Thus, it is impossible to learn remote dependencies, which is not conducive to the extraction of global semantic information that is critical to intensive tasks, especially the semantic segmentation task. To address this limitation, some recent methods propose the use of large convolutional kernel~\cite{peng2017large}, dilated convolution~\cite{chen2017deeplab}, and feature pyramids~\cite{zhao2017pyramid} to expand the sensory field. Another approach is to integrate Non-local~\cite{wang2018non} from the natural language processing (NLP)~\cite{chowdhary2020natural} domains into the FCN structure, which is designed to model the global interaction of all pixels in the feature map, but with high memory and high computational costs. On the other hand, researchers began experimenting with completely removing convolution and exploring a model that used only attention modules alone, Transformer~\cite{vaswani2017attention}, which was designed to model sequence-to-sequence long-range dependencies and capture relationships anywhere in the sequence. 

Unlike previous CNN-based approaches, Transformer~\cite{vaswani2017attention} is not only powerful in terms of global context modeling but also achieves good results on downstream tasks in the case of large-scale pre-training. In ~\cite{dosovitskiy2020image}, a visual Transformer (ViT) is proposed to perform image recognition tasks using a two-dimensional image block with positional embedding as input. However, the disadvantage of ViT~\cite{dosovitskiy2020image} compared to CNN is that it must be pre-trained on large data sets. The image resolution is much larger than words in NLP~\cite{chowdhary2020natural}. Many computer vision tasks, such as semantic segmentation, require intensive prediction at the pixel level, which is difficult to process for Transformers on high-resolution images because its computational complexity of self-attention is related to the size of the image at the quadratic level. In addition, when Transformer is used in the image processing field, the two-dimensional image is sliced and fed into the model as a one-dimensional sequence, thus breaking the connection between local structures and focusing only on the global context at all stages. As a result, low-resolution features lack detailed localization information that cannot be effectively recovered by directly upsampling to full resolution, resulting in rough segmentation results.

Although Transformer can achieve global information modeling, it cannot extract fine spatial details. On the contrary, CNN can provide a way to extract low-level visual cues that compensate well for this fine spatial detail. Therefore, some methods try to combine CNN with Transformer to handle semantic segmentation tasks. For example, in the field of medical image segmentation, TransUNet~\cite{chen2021transunet}, TransBTS~\cite{wang2021transbts}, and TransFuse~\cite{zhang2021transfuse} have achieved satisfactory results. Inspired by this, we also propose a lightweight real-time semantic segmentation model, named LETNet, based on the CNN and Transformer. As depicted in Fig.~\ref{Figure1}, our LETNet achieves a good balance between the performance, model size, and inference speed of the model. The main contribution of this paper is three folds:
\begin{itemize}
\item We propose a Lightweight Dilated Bottleneck (LDB) to extract important semantic information. LDB consists of dilated convolution and depth-wise separable convolution, achieving extreme weight reduction in terms of parameters and computational quantities. 
\item We propose a hybrid network, LETNet, for semantic segmentation. LETNet adopts the most concise encoder-decoder structure and regards the efficient Transformer as a capsule network to learn global information. Meanwhile, a Feature Enhancement (FM) module is added to the jump connection to help supplement the boundary detail information when restoring the resolution. 
\item LETNet achieved 72.8\% mIoU on the cityscapes test set on the single RTX3090 hardware platform with only 0.95M of parameter quantities and 70.5\% of the good performance on the CamVid dataset. The performance is better than most existing models.
\end{itemize}

\begin{figure*}[htbp]
	\centerline{\includegraphics[width=18cm]{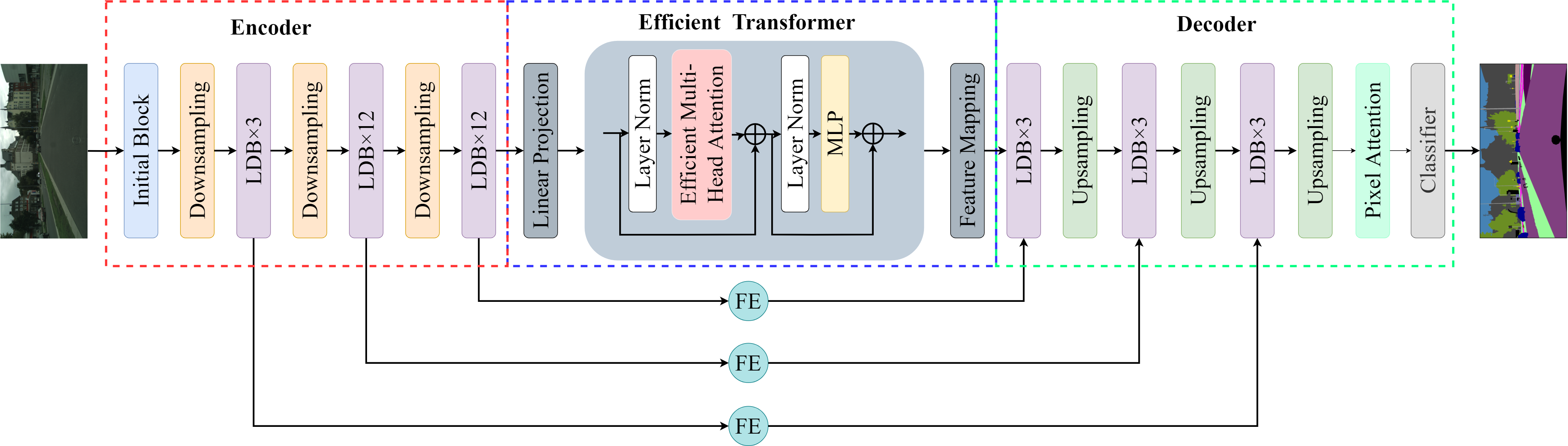}}
	\caption{The proposed Lightweight Real-time Semantic Segmentation Network with Efficient Transformer and CNN (LETNet).}
	\label{Figure2}
\end{figure*}

\section{Related Work}
\subsection{CNN-based Semantic Segmentation Methods}

Owing to the powerful feature representation capabilities of convolutional neural networks, semantic segmentation methods have also made great progress~\cite{gao2021mscfnet}. The groundbreaking article based on CNN was FCN~\cite{long2015fully}, after which many architectures have been refined on this basis. To alleviate the contradiction between image resolution and limited receptive field, DeepLab~\cite{chen2017deeplab} and PSPNet~\cite{zhao2017pyramid} employed parallel atrous convolutions to build an atrous spatial pyramid pooling (ASPP) module, which introduces good descriptors for various scale contextual information. Additionally, with the advantages of modeling feature dependencies, the self-attention mechanism has attracted the interest of many scholars. For instance, 
Based on SENet~\cite{hu2018squeeze}, a local cross-channel interaction strategy without dimensionality reduction and a method for adaptively selecting the size of one-dimensional convolution kernels are proposed in ECANet~\cite{wang2020eca}. In addition, there is an attention mechanism commonly used in NLP~\cite{chowdhary2020natural} to model long-distance dependencies. Typical of these is Non-local neural networks~\cite{wang2018non}, which uses the similarity of two points to weight the features of each position. DANet~\cite{fu2019dual} used ResNet~\cite{he2016deep} as the backbone network, followed by an attention module composed of spatial dimension and channel dimension in parallel for capturing long-range deep features dependencies to improve the segmentation result. CCNet~\cite{huang2019ccnet} was improved to calculate the association between the pixel and all the pixels in the row and column, which economizes the computational burden. LRNNet~\cite{jiang2020lrnnet} proposed an effective simplified Non-local module that uses regional singular vectors to generate more simplified and representative features to model remote dependency and global feature selection. 

Although these types of methods achieve good results, they do not change the fact that non-local is essentially a pixel-wise matrix algorithm, which still makes the computer face a huge computational challenge.
So the lightweight network came into being. For example, ICNet~\cite{zhao2018icnet} used multi-scale images as input where high-level semantic information and low-level spatial details are utilized. BiseNet~\cite{yu2018bisenet} and BiseNet-v2~\cite{yu2021bisenet} proposed two-path architecture, one branch is responsible for extracting deep semantic information, and the other high-resolution shallow branch is responsible for providing detailed information supplement. 
DFANet~\cite{li2019dfanet} utilized a feature reuse policy, which enhanced the interaction and aggregation of features at different levels. Furthermore, point-wise attention is used at the end of each stage to enhance the feature expression ability while ensuring that the computation is small. ESPNet~\cite{mehta2018espnet} and ESPNet-v2~\cite{mehta2019espnetv2} reduced the number of parameters and computation by integrating decomposed convolution into point-wise convolution and dilation convolution. In addition, NRD~\cite{zhang2021dynamic} used dynamic convolutional neural networks to extract feature information from images. However, CNN-based methods always have a problem that cannot be completely solved, and that is the limitation exhibited by modeling long-range relationships. While existing methods only resort to building a deep encoder and downsampling operations, the negative effects are redundant parameters and the loss of more local details.

\subsection{Transformer-based Semantic Segmentation Methods}
Transformer was first proposed in~\cite{vaswani2017attention} and has achieved great success in natural language processing. Unlike CNN, Transformer is not only powerful in terms of global context modeling but also achieves good results on downstream tasks in the case of large-scale pre-training. For example, ViT~\cite{dosovitskiy2020image} proposed to perform image recognition tasks with 2D image patches with position embeddings as input. DETR~\cite{zhu2020deformable} and the deformable version utilized Transformers Encoder-decoder to fuse context in the detection head. SETR~\cite{zheng2021rethinking} solved semantic segmentation from the perspective of sequence to sequence. It abandoned CNN and is a structure completely based on Transformer. Meanwhile, SegFormer~\cite{xie2021segformer}  proposed a hierarchical encoder structure, output multi-scale features, and fuse them in the decoder. After that, some deformed structures were proposed for medical image segmentation~\cite{chen2021transunet, wang2021transbts, zhang2021transfuse, cao2021swin}.

Although the above methods have achieved good results, since the computational complexity of the Transformer is proportional to the square of the image size, it will increase a lot of computational burdens. In addition, when using Transformer in the image domain, the input patches are regarded as a one-dimensional sequence input to the model, which destroys the connection between local structures and only focuses on the global context modeling at the stage. The lack of detailed localization information leads to coarse segmentation results. On the other hand, the CNN architecture provides a path to extract low-level vision cues that can compensate well for such fine spatial details. Therefore, we aim to explore a method combining CNN with Transformer to handle the segmentation task and use an efficient Transformer to reduce memory consumption. 

\begin{figure}[t]
	\centerline{\includegraphics[width=8cm]{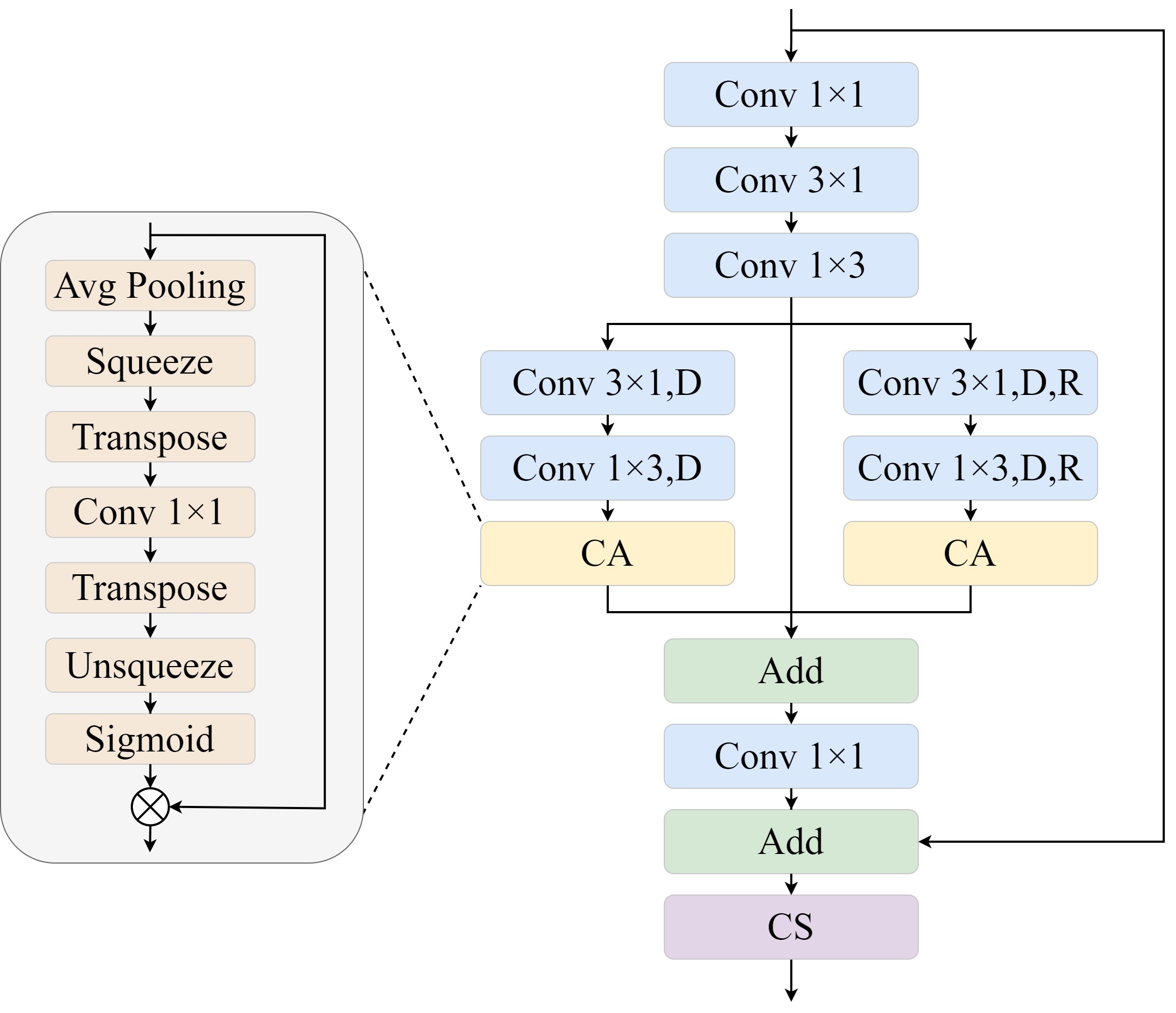}}
	\caption{The proposed Lightweight Dilated Bottleneck (LDB). Among them, $D$ represents depth-wise convolution, $R$ is the kernel of dilated convolution, $CA$ means Channel Attention, and $CS$ denotes the channel shuffle operation.}
	\label{Figure3}
\end{figure}

\section{Proposed Method}
\subsection{Network Architecture}\label{AA}
As shown in Fig.~\ref{Figure2}, LETNet comprises an Encoder, an Decoder, an Efficient Transformer, and three long skip connections. Specifically, the Encoder and Decoder are CNN structures used to extract local features for better image representation. The transformer can reflect complex spatial transformation and long-distance feature dependencies by self-attention and multi-layer perceptron (MLP) structure to obtain global feature representation. The three long-distance connections are inspired by UNet~\cite{ronneberger2015u}, which combines low-level spatial information with high-level semantic information for high-quality segmentation. 

\subsection{Lightweight Dilated Bottleneck (LDB)}
As shown in Fig.~\ref{Figure3}, the structure of LDB adopted the idea of ResNet~\cite{he2016deep} on the whole, and the module is designed as a residual module to collect more feature information while the number of network layers is as small as possible. Specifically, at the bottleneck, the number of channels of the input feature is reduced to half by $1\times1$ convolution. After reducing the number of channels, the amount of parameters and calculations is greatly reduced. Although this will lose a part of the accuracy, it will be more beneficial to stack two modules more than make up for the loss at this point. At the same time, due to the use of $1\times1$ convolution, the network depth must be deepened to obtain a larger receptive field. Therefore, after the $1\times1$ convolution, the decomposed convolutions of $3\times1$ and $1\times3$ are added to expand the feeling to capture a wider range of contextual information. Moreover, decomposed convolution is also based on considering the number of parameters and the amount of calculation. Similarly, in the next two-branch structure, both branches also use decompose convolution, one of which is responsible for local and short-distance feature information, and the other uses atrous convolution, which is responsible for extracting feature information from a larger receptive field under different atrous rates. Next to these two branches are channel attentions, inspired by ECANet~\cite{wang2020eca}, which aims to build an attention matrix in the channel dimension to enhance feature expression and suppress noise interference because, for CNN, most of the feature information is contained in the channel. Then, the two low-dimensional branches and middle features are fused and input to a $1\times1$ point-wise convolution below to restore the number of channels of the feature map to be the same as the number of channels of the input feature map. Finally, the strategy of channel shuffle is used to avoid the drawbacks of information independence and no correlation between channels caused by depth-wise convolution and to promote the exchange of semantic information between different channels. The complete operation is shown as follows:
\begin{equation}
{F_1} = {f_{1 \times 3}}\left( {{f_{3 \times 1}}\left( {{f_{1 \times 1}}\left( {{x}} \right)} \right)} \right),
\end{equation}
\begin{equation}
{F_{21}} = {f_{CA}}\left( {{f_{1 \times 3,D}}\left( {{f_{3 \times 1,D}}\left( {{F_1}} \right)} \right)} \right),
\end{equation}
\begin{equation}
{F_{22}} = {f_{CA}}\left( {{f_{1 \times 3,D,R}}\left( {{f_{3 \times 1,D,R}}\left( {{F_1}} \right)} \right)} \right),
\end{equation}
\begin{equation}
{y} = {f_{CS}}\left( {{f_{1 \times 1}}\left( {{F_{1}} + {F_{21}} + {F_{22}}} \right)} + x \right),
\end{equation}
where $x$ represents the input feature maps, $y$ represents the output feature map, and $f_{k\times k}(\cdot)$ are convolution operation. 


\subsection{Efficient Transformer (ET)}
\begin{figure}
	\centerline{\includegraphics[width=9cm]{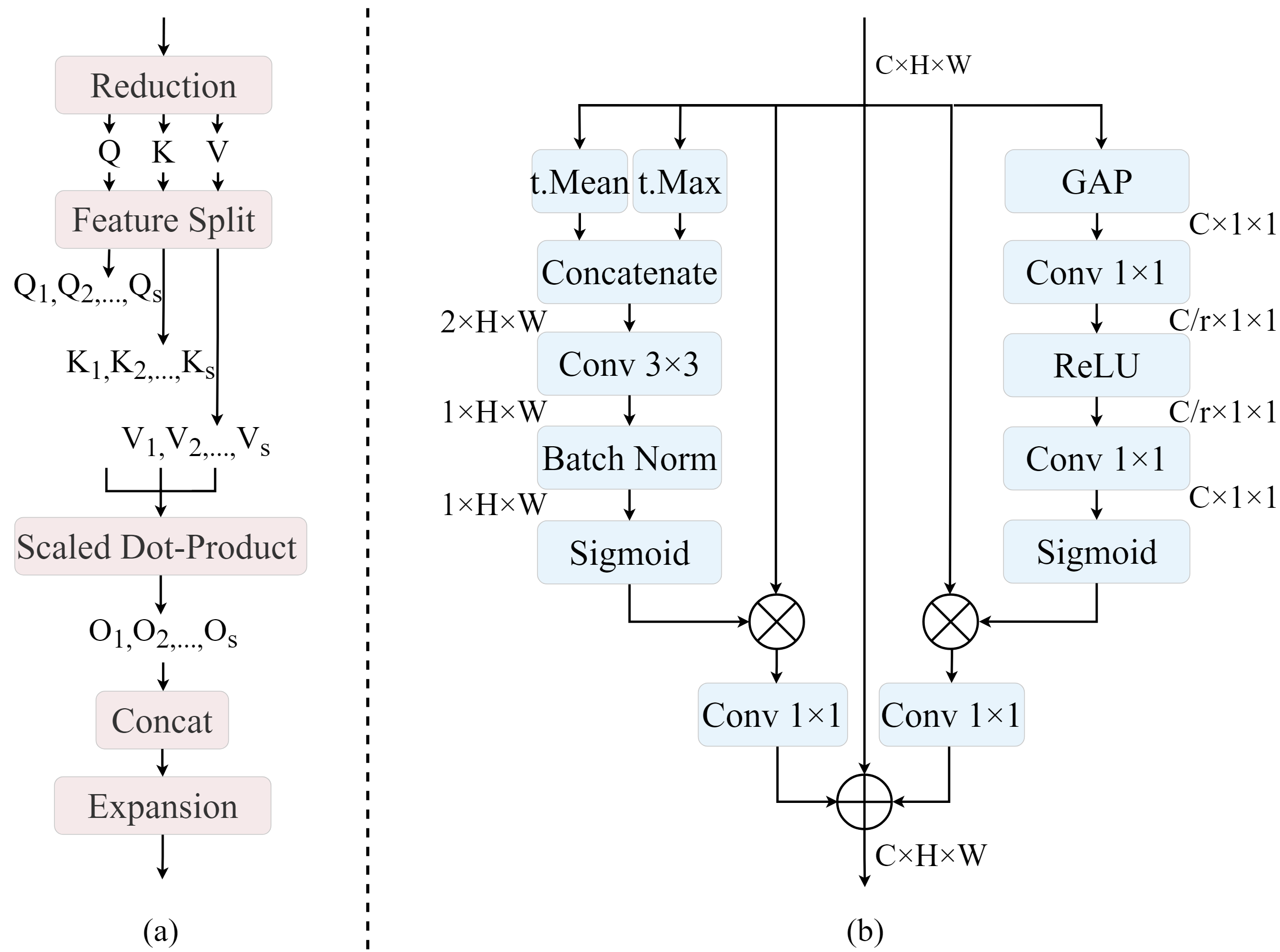}}
	\caption{Schematic diagram of the (a) Efficient Multi-Head Attention (EMHA) and (b) Feature Enhancement (FE) module. Please zoom in for details.}
	\label{Figure4}
\end{figure}

As we mentioned before, despite its advantages in local feature extraction, the ability of CNN to capture global representations is still insufficient, which is important for many high-level computer vision tasks. To deal with this problem, we introduce Transformer to learn long-range dependencies. However, in image processing tasks, since the input image resolution is much larger than the words in the natural language processing field. We introduce the Efficient Transformer (ET), which is inspired by ETSR~\cite{lu2021efficient}. Different from the traditional Transformer, ET occupies fewer computing resources. Meanwhile, to avoid excessive memory usage and computational load, we abandon the series connection of multiple ETs in ETSR~\cite{lu2021efficient} and only use one ET as a capsule network, which is placed in the middle of the entire network. As we all know, Transformer consists of two layer normalizations, one Multi-Head Attention (MHA), and one Multi-Layer Perception (MLP). The biggest difference between ET and the original Transformer is that MHA. After the layer normalization, ET sets up a reduction layer to halve the number of channels, which reduces part of the computation. Then, a linear layer projects the feature map into three matrices, ${Q}$ (query), ${K}$ (key), and ${V}$ (value). Specifically, in EMHA, ${Q}$, ${K}$, and ${V}$ are first split into $s$ segments, and then a scaled dot product attention of ${Q_i}$, ${K_i}$, and ${V_i}$ is executed correspondingly. After that, we concatenate the obtained ${O_1}$.....${O_s}$ to get the whole output $O$, In fact, it relies on the idea of group convolution, splitting large matrices into small matrices and then calculating and finally merging, so as to achieve the purpose of reducing the amount of calculation. Finally, the expansion layer is employed to restore the number of channels. The architecture of EMHA is shown in Fig.~\ref {Figure4} (a) and the Scaled Dot-Product Attention (SDPA) operation can be defined as:
\begin{equation}
{O_i} = soft\max \left( {\frac{{{Q_i}K_i^T}}{{\sqrt d }}} \right){V_i},
\end{equation}
where $Q$, $K$, and $V$ represent $query$, $key$, and $value$ metrices, $d$ is the embedding dimension. Afterward, all the outputs ($O_{1}, O_{2}, ..., O_{s}$) of SDPA are concatenated together to generate the whole output feature $O$.





\subsection{Feature Enhancement (FE)}
In the neural network, the lower layer has high resolution and accurate spatial information (the resolution corresponds to the spatial position) but has little semantic information. In contrast, the high layer has low resolution and lacks spatial position information but rich semantic information. Therefore, in the segmentation task, to make the high-level information also have enough spatial information, the low-level spatial information, and high-level semantic information are usually combined to perform high-quality segmentation. Therefore, we use the UNet-style structure to fuse the high-level and low-level feature maps of the same resolution. At the same time, in the process of three long connections, we propose a Feature Enhancement (FE) module to improve the ability of feature expression. As shown in the fig.~\ref{Figure4} (b), feature dependency modeling is carried out from two dimensions, one is the channel dimension, the other is the spatial dimension, and the two dimensions are transformed at the same time and finally fused to transmit the low-level information to the high-level more effectively. The operation can be defined as:
\begin{equation}
{M_C} = X * \sigma \left( {f_{1 \times 1}^C\left( {\gamma \left( {f_{1 \times 1}^{\frac{C}{r}}\left( {{f_{AP}}\left( X \right)} \right)} \right)} \right)} \right),
\end{equation}
\begin{equation}
{M_S} = X * \sigma \left( {{\rm B}\left( {{f_{3 \times 3}}\left( {Concat\left[ {{f_{AP}}\left( X \right),{f_{MP}}\left( X \right)} \right]} \right)} \right)} \right),
\end{equation}
\begin{equation}
Y = {f_{1 \times 1}}\left( {{M_C}} \right) + {f_{1 \times 1}}\left( {{M_S}} \right) + X,
\end{equation}
where $X$ is the input feature, ${M_C}$ represents the output of channel dimension, ${M_S}$ represents the output of spatial dimension, ${\sigma}$ denotes the sigmoid function, ${\gamma}$ means ReLU function, $B$ represents Batch Normalization, $C$ is the channel of the feature map, $r$ means reduction, $f_{AP}(\cdot)$ and $f_{MP}(\cdot)$ denote the average pooling and max pooling operations, respectively.


\begin{figure}
	\centerline{\includegraphics[width=8cm]{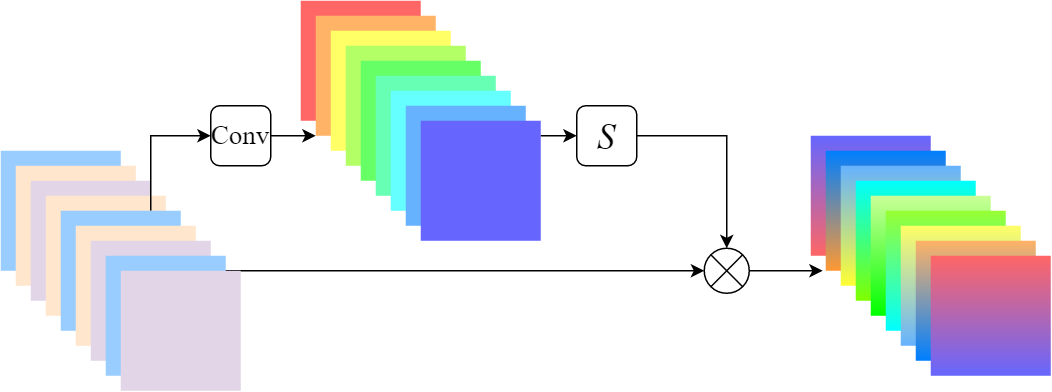}}
	\caption{Schematic diagram of the pixel attention mechanism. Among them, $S$ means the $Sigmoid$ function.}
	\label{Figure6}
\end{figure}

\subsection{Pixel Attention (PA)}
The pixel attention (PA) mechanism learns weights based on the importance of features at different pixel positions. This means that each channel has the same weight but the weights are different at different pixel positions for the same channel. The pixel attention pays more attention to the edges and textures of objects in the image, so adding PA can facilitate the recovery of edge detail information, thereby improving the performance of segmentation. The operation is shown in Fig.~\ref{Figure6} and the formula is as follows:
\begin{equation}
y = \sigma\left({{f_{1 \times 1}}\left(X \right)} \right)*X,
\end{equation}
where $X$ is the input feature, ${\sigma}$ denotes the sigmoid function, and $f_{1 \times 1}(\cdot)$ is the convolutional layer with kernel size of $1$.

\section{Experiments}
\subsection{Datasets}
\textbf{Cityscapes}: The resolution of images in this dataset is 2048$\times$1024, collected from German and French urban road scenes in 50 different cities, including pedestrians, roads, vehicles, etc. It has 19 categories for the evaluation of semantic segmentation. Among them, 5000 finely annotated images are further divided into 2075, 500, and 1525 for training, validation, and testing, respectively.

\textbf{Camvid}: It contains 701 urban road images of 960$\times$720, which have 11 categories, and the finely annotated images are divided into 367, 101, and 233 for training, validation, and testing, respectively.

\subsection{Model Settings}
In this work, we use the PyTorch framework to build the model and train it on an RTX3090 GPU. In Table~\ref{Table1}, we show the detail of the model settings of LETNet on the Cityscapes and CamVid datasets. Meanwhile, the learning rate varies with iterations and can be calculated as follows:
\begin{equation}
lr = l{r_{initial}} \times {\left( {1 - \frac{{iteration}}{{\max \_iteration}}} \right)^{0.9}},
\end{equation}
where $l{r_{initial}}$ represents the initial learning rate. It is worth noting that we train Cityscapes and CamVid separately with different parameter settings since the resolution of the datasets is different.

\begin{table}[t]
	\caption{The detail of the model settings.}
	\begin{center}
		\setlength\tabcolsep{6pt} 
		
		\begin{tabular}{c|c|c}
			\toprule
			Dataset                 & Cityscapes          						& CamVid    \\ \hline \midrule
			Batch size              & 6                   						& 8 \\ \midrule
			Loss function           & \multicolumn{2}{c}{CrossEntropy Loss} \\ \midrule
			Optimization method     & SGD(momentum 0.9)             		& Adam(momentum 0.9)  \\ \midrule
			Weight decay            & $1 \times {10^{{\rm{ - }}4}}$    	& $2 \times {10^{{\rm{ - }}4}}$ \\ \midrule
			Initial learning rate   & $4.5 \times {10^{{\rm{ - }}2}}$    	& $1 \times {10^{{\rm{ - }}3}}$ \\ \midrule
			Learning rate policy    & \multicolumn{2}{c}{Poly}  \\ \bottomrule
		\end{tabular}
		\label{Table1}
	\end{center}
\end{table}

\begin{figure}[t]
 \centerline{\includegraphics[width=8.8cm]{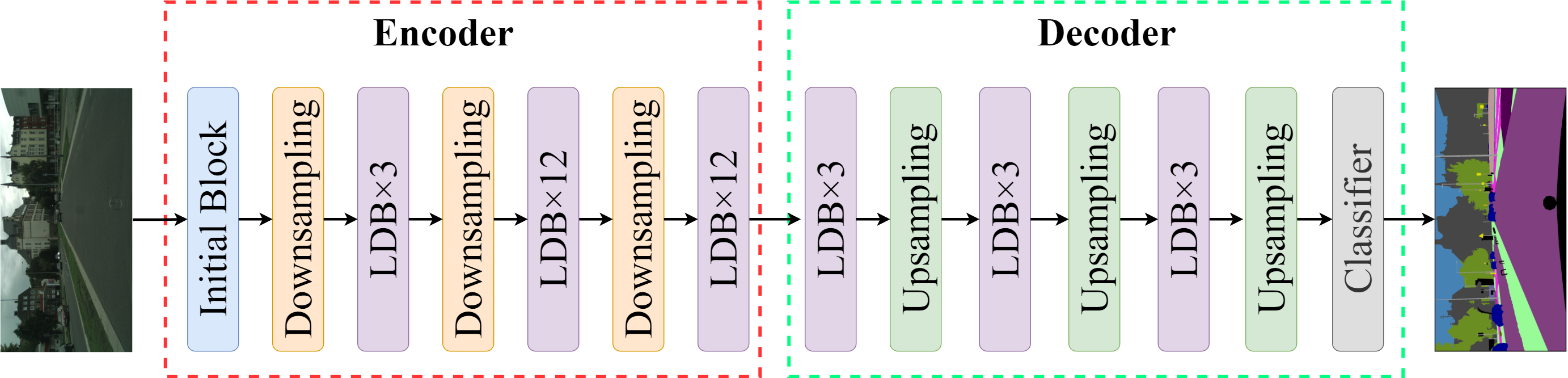}}
 \caption{The architecture of the baseline model.}
 \label{Figure7}
\end{figure}

\begin{table*}[htbp]
\caption{Ablation study for the proposed modules 
on Cityscapes. L1, L2, L3: Line1, Line2, Line3, FE: Feature Enhancement, $*$ represents the final version, and the Difference means the gap in performance between the model and baseline in each group.}
	\begin{center}
		\setlength\tabcolsep{8pt}
		\begin{tabular}{@{}l|lcccc@{}}
			\toprule
\rule{0pt}{8pt}
			\multirow{1}{*}{}
			&Method         & Parameter (K)$\downarrow$     & FLOPs (G)$\downarrow$     & mIoU (\%)$\uparrow$  & Difference  \\ \midrule\hline
\rule{0pt}{8pt}
			\multirow{5}{*}{\textbf{A:} Long Connection} 
\rule{0pt}{8pt}
			&Baseline           											              & 723,400 & 13.068300560 & 69.85 & -\\
\rule{0pt}{8pt}
			&\textbf{A1:} Baseline$+$L1      											& 731,128  & 13.319958800 & 70.43 & $+$0.58\\
\rule{0pt}{8pt}
			&\textbf{A2:} Baseline$+$L1$+$L2 											& 758,856  & 13.546451216 & 70.79 & $+$0.94\\
\rule{0pt}{8pt}
			&\textbf{A3:} Baseline$+$L1$+$L2$+$L3   									& 761,976  & 13.552742672 & 71.01 & $+$1.16\\
\midrule

\rule{0pt}{8pt}
			\multirow{4}{*}{\textbf{B:} Feature Enhancement}
\rule{0pt}{8pt}
			&Baseline           												           & 723,400  & 13.068300560 & 69.85 & -\\
\rule{0pt}{8pt}
			&\textbf{B1:} Baseline$+$L1(FE)    										& 731,229 & 13.323170112 & 70.63 & $+$0.78 \\
\rule{0pt}{8pt}
			&\textbf{B2:} Baseline$+$L1(FE)$+$L2(FE)  								& 759,058 & 13.550465392 & 71.14 & $+$1.29 \\
\rule{0pt}{8pt}
			&\textbf{B3:} Baseline$+$L1(FE)$+$L2(FE)$+$L3(FE)   					& 762,279  & 13.556957648 & 71.46 & $+$1.61 \\ 
			
\midrule
\rule{0pt}{8pt}
\multirow{4}{*}{\textbf{C:} Efficient Transformer}
\rule{0pt}{8pt}
			&Baseline              											            & 723,400 & 13.068300560 & 69.85 & - \\
\rule{0pt}{8pt}
			&\textbf{C1:} Baseline$+$ET         										 & 911,824 & 13.068487760 & 72.92 & $+$3.07 \\
\rule{0pt}{8pt}
			&\textbf{C2:} Baseline$+$ET$+$L1(FE)$+$L2(FE)$+$L3(FE)            & 950,703  & 13.557144848 & 74.53 & $+$4.68\\
\rule{0pt}{8pt}
			&\cellcolor{gray!20}\textbf{C3*:} Baseline$+$ET$+$L1(FE)$+$L2(FE)$+$L3(FE)$+$PA   &\cellcolor{gray!20}950,975 &\cellcolor{gray!20}13.590699280 &\cellcolor{gray!20}75.00 &\cellcolor{gray!20}$+$5.15\\

 \bottomrule
		\end{tabular}
		\label{Table2}
	\end{center}
\end{table*}

\begin{table*}[htbp]
	\caption{Comparisons with the state-of-arts methods on the Cityscapes dataset.}
	\begin{center}
		\setlength\tabcolsep{9pt}
		\begin{tabular}{@{}c|lccccccc@{}}
			\toprule
			\rule{0pt}{8pt}
			\multirow{1}{*}{}
				& Methods  & Year & Resolution & Backbone & Parameter (M)$\downarrow$ & FLOPs (G)$\downarrow$ & Speed (FPS)$\uparrow$ & mIoU (\%)$\uparrow$  \\ \hline \midrule
			\rule{0pt}{8pt}
			\multirow{10}{*}{\rotatebox{90}{Large Size}} 
				& DeepLab~\cite{chen2017deeplab} 				& 2015 & 512$\times$1024  & ResNet-101 & 262.10  & 457.8   & 0.25    & 63.5 \\ 
			\rule{0pt}{8pt}
				& DeepLab-v3+~\cite{chen2018encoder}  			& 2018 & -         & Xception   & \cellcolor{gray!10}15.40   & 555.4   & \cellcolor{gray!30}8.40    & 75.2 \\ 
			\rule{0pt}{8pt}
				& DenseASPP~\cite{yang2018denseaspp}   			& 2018 & 512$\times$512   & DenseNet   & 35.70   & 632.9   & -       & 80.6 \\
			\rule{0pt}{8pt} 
				& PSPNet~\cite{zhao2017pyramid}      			& 2017 & 713$\times$713   & ResNet-101 & 250.80  & \cellcolor{gray!20}412.2   & 0.78    & 81.2 \\ 
			\rule{0pt}{8pt}
				& DANet~\cite{fu2019dual}      			& 2019 & 1024$\times$1024 & ResNet-101 & 66.60   & 1298.8  & 4.00    & 81.5 \\ 
			\rule{0pt}{8pt}
				& CCNet~\cite{huang2019ccnet}       			& 2019 & 1024$\times$1024 & ResNet-101 & 66.50   & 1153.9  & 4.70    & 81.9 \\ 
			\rule{0pt}{8pt}
				& SETR-PUP~\cite{zheng2021rethinking}    			& 2021 & 768$\times$768   & ViT-Large  & 318.30  & -  		  & 0.50    & 82.2 \\ 
			\rule{0pt}{8pt}
				& SegFormer~\cite{xie2021segformer}   			& 2021 & 1024$\times$2048 & MiT-B5     & 84.70   & 1447.6  & 2.50    & 84.0 \\ 
			\rule{0pt}{8pt}
				& Lawin Transformer~\cite{yan2022lawin} 	& 2022 & 1024$\times$1024 & Swin-L     & -      & 1797    & -       & \cellcolor{gray!50}84.4 \\ \midrule

				\rule{0pt}{8pt}
				\multirow{9}{*}{\rotatebox{90}{Medium Size}}
				& SegNet~\cite{badrinarayanan2017segnet}           & 2017  & 640$\times$360    & VGG-16      & 29.50   & 286.0  & 17   & 57.0 \\
				\rule{0pt}{8pt}
           		& SQNet~\cite{treml2016speeding}            & 2016  & 1024$\times$2048  & SqueezeNet  & -       & 270.0 	& 17	 & 59.8 \\ 
				\rule{0pt}{8pt}
				& BiseNet-v1~\cite{yu2018bisenet}       & 2018  & 768$\times$1536   & Xception    & \cellcolor{gray!10}5.80		& 14.8 	& \cellcolor{gray!30}106	 & 68.4 \\
				\rule{0pt}{8pt}
				& ICNet~\cite{zhao2018icnet}            & 2018  & 1024$\times$2048  & PSPNet-50   & 26.50		& 28.3 	& 30	 & 69.5 \\
				\rule{0pt}{8pt}
				& DFANet~\cite{li2019dfanet}           & 2019  & 1024$\times$1024  & Xception    & 7.80		& \cellcolor{gray!20}3.4 	& 100	 & 71.3 \\
				\rule{0pt}{8pt}
				& STDC1-50~\cite{fan2021rethinking}        & 2021  & 512$\times$1024   & -       	 & 8.40		& - 		& 87   & 71.9 \\
				\rule{0pt}{8pt}
				& FPANet~\cite{wu2022fpanet}           & 2022  & 512$\times$1024   & -       	 & 14.10		& - 		& -    & 72.0 \\
				\rule{0pt}{8pt}
				& HSB-Net~\cite{li2021hierarchical}        & 2021  & 512$\times$1024  & ResNet-34  & 12.10	& - 	& 124   	 & 73.1 \\ 
				\rule{0pt}{8pt}
				& LBN-AA~\cite{dong2020real}          & 2021  & 448$\times$896  & No    & 6.20	& 49.5 	& 51  & \cellcolor{gray!50}73.6 \\ \midrule

				\rule{0pt}{8pt}
				\multirow{16}{*}{\rotatebox{90}{Small Size}}
				& ENet~\cite{paszke2016enet}           & 2016  & 512$\times$1024   & No        & \cellcolor{gray!10}0.36    & 3.8   & 135   & 58.3 \\
				\rule{0pt}{8pt}
           		& ESPNet~\cite{mehta2018espnet}         & 2018  & 512$\times$1024  & ESPNet    & \cellcolor{gray!10}0.36    & - 	  & 113	 & 60.3 \\ 
				\rule{0pt}{8pt}
				& CGNet~\cite{wu2020cgnet}          & 2020  & 360$\times$640    & No    	   & 0.50		& 6.0 	& -	 & 64.8 \\
				\rule{0pt}{8pt}
				& NDNet~\cite{yang2020ndnet}          & 2021  & 1024$\times$2048  & No        & 0.50		& 14.0 	& 40	 & 65.3 \\
				\rule{0pt}{8pt}
				& ESPNet-v2~\cite{mehta2019espnetv2}      & 2019  & 512$\times$1024   & ESPNet-v2 & -		   & \cellcolor{gray!20}2.7 	& 80	 & 66.2 \\
				\rule{0pt}{8pt}
				& ADSCNet~\cite{wang2020adscnet}        & 2020  & 512$\times$1024   & No        & -		   & - 		& 77   & 67.5 \\
				\rule{0pt}{8pt}
				& ERFNet~\cite{romera2017erfnet}         & 2017  & 512$\times$1024   & No        & 2.10		& - 		& 42    & 68.0 \\
				\rule{0pt}{8pt}
				& CFPNet~\cite{lou2021cfpnet}         & 2021  & 1045$\times$2048   & No        & 0.55		& - 	   & 30   & 70.1 \\
				\rule{0pt}{8pt}
				& FPENet~\cite{liu2019feature}         & 2019  & 512$\times$1024   & No        & 0.40		& 12.8 	& 55   & 70.1 \\
				\rule{0pt}{8pt}
				& LEDNet~\cite{wang2019lednet}         & 2019  & 512$\times$1024   & No        & 0.94		& - 	  & 40   & 70.6 \\
				\rule{0pt}{8pt}
				& SGCPNet~\cite{hao2022real}        & 2022  & 1024$\times$2048  & No        & 0.61	   & 4.5 	& 103	 & 70.9 \\ 
				\rule{0pt}{8pt}
				& FBSNet~\cite{gao2022fbsnet}        & 2022  & 512$\times$1024  & No        & 0.62	   		& 9.7 	& 90   	 & 70.9 \\ 
				\rule{0pt}{8pt}
				& EdgeNet~\cite{han2020using}        & 2021  & 512$\times$1024  & No        & -	   		& - 	& 31   	 & 71.0 \\ 
				\rule{0pt}{8pt}
				& MSCFNet~\cite{gao2021mscfnet}        & 2022  & 512$\times$1024  & No        & 1.15	   		& 17.1 	& 50   	 & 71.9 \\ 
				\rule{0pt}{8pt}
				& BiseNet-v2~\cite{yu2021bisenet}       & 2021  & 512$\times$1024   & Xception    & 3.40		& 21.2 	& \cellcolor{gray!30}156	 &  72.6\\
				\rule{0pt}{8pt}
				& MGSeg~\cite{he2021mgseg}          & 2021  & 1024$\times$1024  & ShuffleNet-v2  & 4.50	& 16.2 	& 101   	 & 72.7 \\ 
				
\midrule
				\rule{0pt}{8pt}
				& LETNet (ours) & -  & 512$\times$1024 & No    & 0.95	 & 13.6 & 150  & \cellcolor{gray!50}72.8 \\ 

 \bottomrule
		\end{tabular}
		\label{Table3}
	\end{center}
\end{table*}

\begin{table*}[!t]
\caption{Per-class IoU (\%) results on the Cityscapes test set. “Avg” represents the average results of all these categories. Obviously, our FBSNet achieves the best mIoU results.}
\small
\setlength{\tabcolsep}{2.35pt}
\begin{tabular}{@{}l|c||ccccccccccccccccccc@{}}
\toprule

Methods  & Avg & Bic  & Bus  & Bui  & Car  & Fen  & Mot  & Pol  & Per  & Rid  & Roa  & Sid  & Sky  & Tru  & Tra  & TLi  & Ter  & TSi  & Veg & Wal  \\  \hline\midrule
\rule{0pt}{8pt}
SegNet~\cite{badrinarayanan2017segnet}     & 57.0 & 51.9 & 43.1 & 84.0 & 89.3 & 29.0 & 35.8 & 35.1 & 62.8 & 42.8 & 96.4 & 73.2 & 91.8 & 38.1 & 44.1 & 39.8 & 63.8 & 45.1 & 87.0 & 28.4 \\
\rule{0pt}{8pt}
ENet~\cite{paszke2016enet}       & 58.3 & 55.4 & 50.5 & 75.0 & 90.6 & 33.2 & 38.8 & 43.4 & 65.5 & 38.4 & 96.3 & 74.2 & 90.6 & 36.9 & 48.1 & 34.1 & 61.4 & 44.0 & 88.6 & 32.2 \\
\rule{0pt}{8pt}
ESPNet~\cite{mehta2018espnet}     & 60.3 & 57.2 & 52.5 & 76.2 & 92.3 & 36.1 & 41.8 & 45.0 & 67.0 & 40.9 & 97.0 & 77.5 & 92.6 & 38.1 & 50.1 & 35.6 & 63.2 & 46.3 & 90.8 & 35.0 \\
\rule{0pt}{8pt}
ESPNet-v2~\cite{mehta2019espnetv2}  & 66.2 & 59.9 & 65.9 & 88.8 & 91.8 & 42.1 & 44.2 & 49.3 & 72.9 & 53.1 & 97.3 & 78.6 & 93.3 & 53.0 & 53.2 & 52.6 & 66.8 & 60.0 & 90.5 & 43.5 \\
\rule{0pt}{8pt}
ICNet~\cite{zhao2018icnet}      & 69.5 & 70.5 &\cellcolor{gray!30} 72.7 & 89.7 & 92.6 & 48.9 & 53.6 & 61.5 & 74.6 & 56.1 & 97.1 & 79.2 & 93.5 & 51.3 & 51.3 & 60.4 & 68.3 & 63.4 & 91.5 & 43.2 \\

\rule{0pt}{8pt}
LEDNet~\cite{wang2019lednet}     & 70.6 &\cellcolor{gray!30}71.6 & 64.0 & 91.6 & 90.9 & 49.9 & 44.4 &\cellcolor{gray!30} 62.8 & 76.2 & 53.7 & 98.1 & 79.5 &94.9 &\cellcolor{gray!30} 64.4 & 52.7 & 61.3 & 61.2 &\cellcolor{gray!30} 72.8 & 92.6 & 47.7 \\

\rule{0pt}{8pt}
FBSNet~\cite{gao2022fbsnet}
&70.9 & 70.1 & 56.0 & 91.5 & 93.9 & 53.5 & 56.2 & 62.5 & 82.5 &\cellcolor{gray!30} 63.8 & 98.0 & 83.2 & 94.4 & 50.5 & 37.6 &\cellcolor{gray!30} 67.6 & 70.5 & 71.5 &\cellcolor{gray!30} 92.7 & 50.9
\\
\rule{0pt}{8pt}
EdgeNet~\cite{han2020using}     & 71.0 & 67.7 & 60.9 & 91.6 & 94.3 & 50.6 & 55.3 & 62.6 & 80.4 & 61.1 & 98.1 & 83.1 & 94.9 & 50.0 & 52.5 & 67.2 & 69.7 & 71.4 & 92.4 & 45.4 \\
\rule{0pt}{8pt}
MSCFNet~\cite{gao2021mscfnet}
&71.9 & 70.2 & 66.1 & 91.0 & 94.1 & 52.5 &\cellcolor{gray!30} 57.6 & 61.2 &\cellcolor{gray!30} 82.7 & 62.7 & 97.7 &82.8 & 94.3 & 50.9 & 51.9 & 67.1 & 70.2 & 71.4 & 92.3 & 49.0
\\
\midrule
\rule{0pt}{8pt}

LETNe (ours) & 72.8 & 69.3 & 72.4 &\cellcolor{gray!30} 91.6 &\cellcolor{gray!30}94.4 &\cellcolor{gray!30} 53.7 & 56.1 &61.0 & 82.3 & 61.7 &\cellcolor{gray!30}98.2 &\cellcolor{gray!30} 83.6 & \cellcolor{gray!30}94.9 & 55.0 &\cellcolor{gray!30} 57.0 &66.7 & \cellcolor{gray!30}70.5 & 70.5 & 92.5 &\cellcolor{gray!30} 50.9 \\

\bottomrule
\end{tabular}
\label{Table4}
\end{table*}

\begin{table*}[t]
	\caption{Comparisons with the state-of-art methods on the CamVid dataset}
	\begin{center}
		\setlength\tabcolsep{18pt}
		\begin{tabular}{@{}l|ccccccc@{}}
			\toprule
\rule{0pt}{8pt}
			Method  & Year & Resolution & Backbone & Parameter (M)$\downarrow$  & Speed (FPS)$\uparrow$ & mIoU (\%)$\uparrow$  \\ \hline \midrule
\rule{0pt}{8pt}
			ENet~\cite{paszke2016enet}  & 2016  & 360$\times$480  & No &\cellcolor{gray!10} 0.36  & 61  & 51.3   \\
\rule{0pt}{8pt}
			SegNet~\cite{badrinarayanan2017segnet}    & 2017  & 360$\times$480  & VGG-16 & 29.50  & 29  & 55.6 \\
\rule{0pt}{8pt}
			NDNet~\cite{yang2020ndnet}  & 2021  & 360$\times$480  & - & 0.50  & -  & 57.2\\
\rule{0pt}{8pt}
			DFANet~\cite{li2019dfanet}  & 2019  & 720$\times$960  & Xception & 7.80  & 120  & 64.7\\
\rule{0pt}{8pt}
			BiseNet-v1~\cite{yu2018bisenet}  & 2018  & 720$\times$960  & Xception & 5.80  & 116  & 65.6 \\
\rule{0pt}{8pt}
			DABNet~\cite{li2019dabnet} & 2019   & 360$\times$480 & No    & 0.76 & -   & 66.4 \\
\rule{0pt}{8pt}
         	FDDWNet~\cite{liu2020fddwnet} & 2020  & 360$\times$480  & No & 0.80  & 79  & 66.9 \\
\rule{0pt}{8pt}
			ICNet~\cite{zhao2018icnet}  & 2018  & 720$\times$960  & PSPNet-50 & 26.50  & 28  & 67.1\\
\rule{0pt}{8pt}
			LBN-AA~\cite{dong2020real}  & 2021  & 720$\times$960  & No & 6.20  & 39  & 68.0 \\
\rule{0pt}{8pt}
			BiseNet-v2~\cite{yu2021bisenet}  & 2020  & 720$\times$960  & ResNet & 49.00  & -  & 68.7 \\
\rule{0pt}{8pt}
			FBSNet~\cite{gao2022fbsnet}  & 2022  & 360$\times$480  & No & 0.62  & 120  & 68.9 \\
\rule{0pt}{8pt}
			SGCPNet~\cite{hao2022real}  & 2022  & 720$\times$960  & No & 0.61  &\cellcolor{gray!30} 278  & 69.0 \\
\rule{0pt}{8pt}
			MSCFNet~\cite{gao2021mscfnet}  & 2021  & 360$\times$480  & No & 1.15  &-  & 69.3 \\\midrule
\rule{0pt}{8pt}
			LETNet (ours) & 2022 & 360$\times$480  & No & 0.95  & 200   &\cellcolor{gray!50} 70.5 \\ \bottomrule
		\end{tabular}
		\label{Table5}
	\end{center}
\end{table*}

\begin{figure*}[t]
	\centerline{\includegraphics[width=18cm]{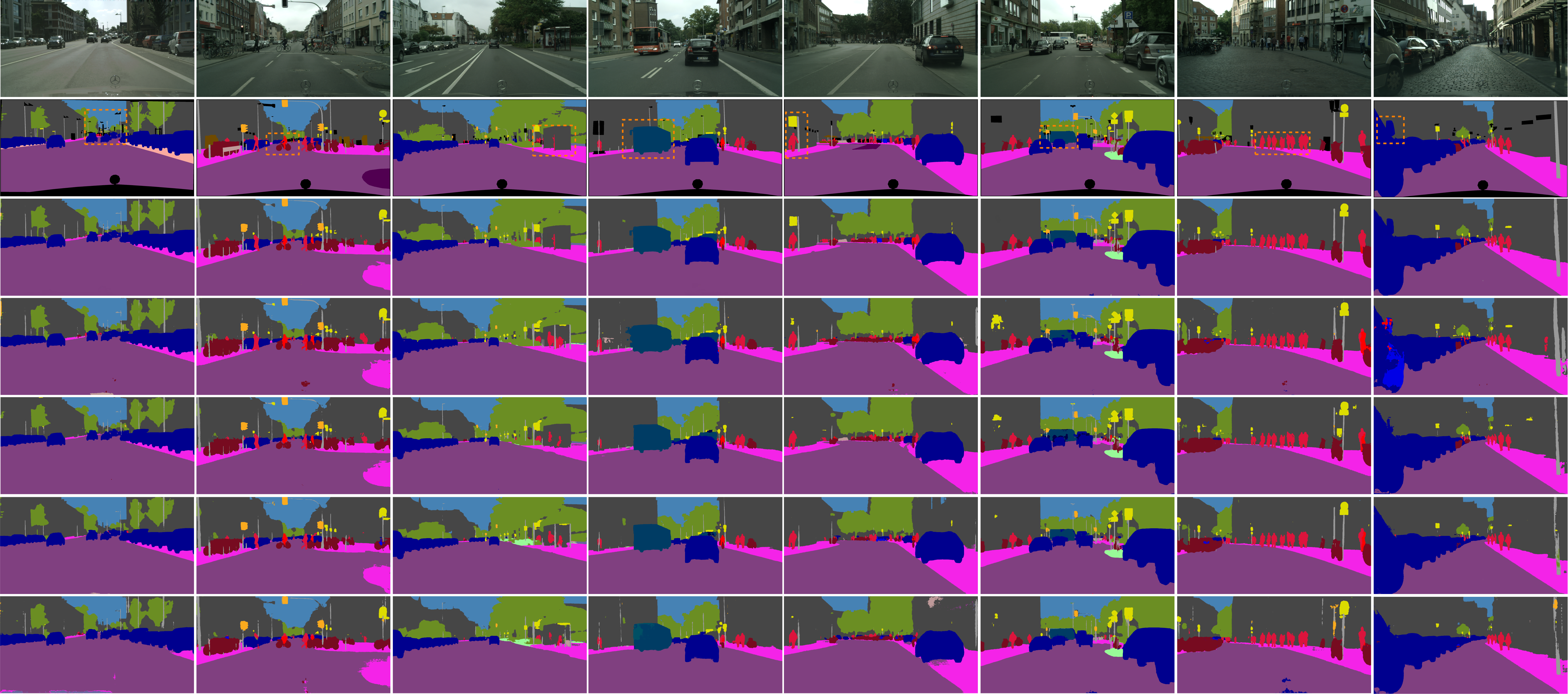}}
	\caption{Visual comparisons on the Cityscapes dataset. From top to bottom are original input images, ground truths, and segmentation results from our \textbf{LETNet}, LEDNet~\cite{wang2019lednet}, ERFNet~\cite{romera2017erfnet}, ESPNet~\cite{mehta2018espnet}, and ENet~\cite{paszke2016enet}.}
	\label{Figure8}
\end{figure*}

\begin{figure*}[t]
	\centerline{\includegraphics[width=18cm]{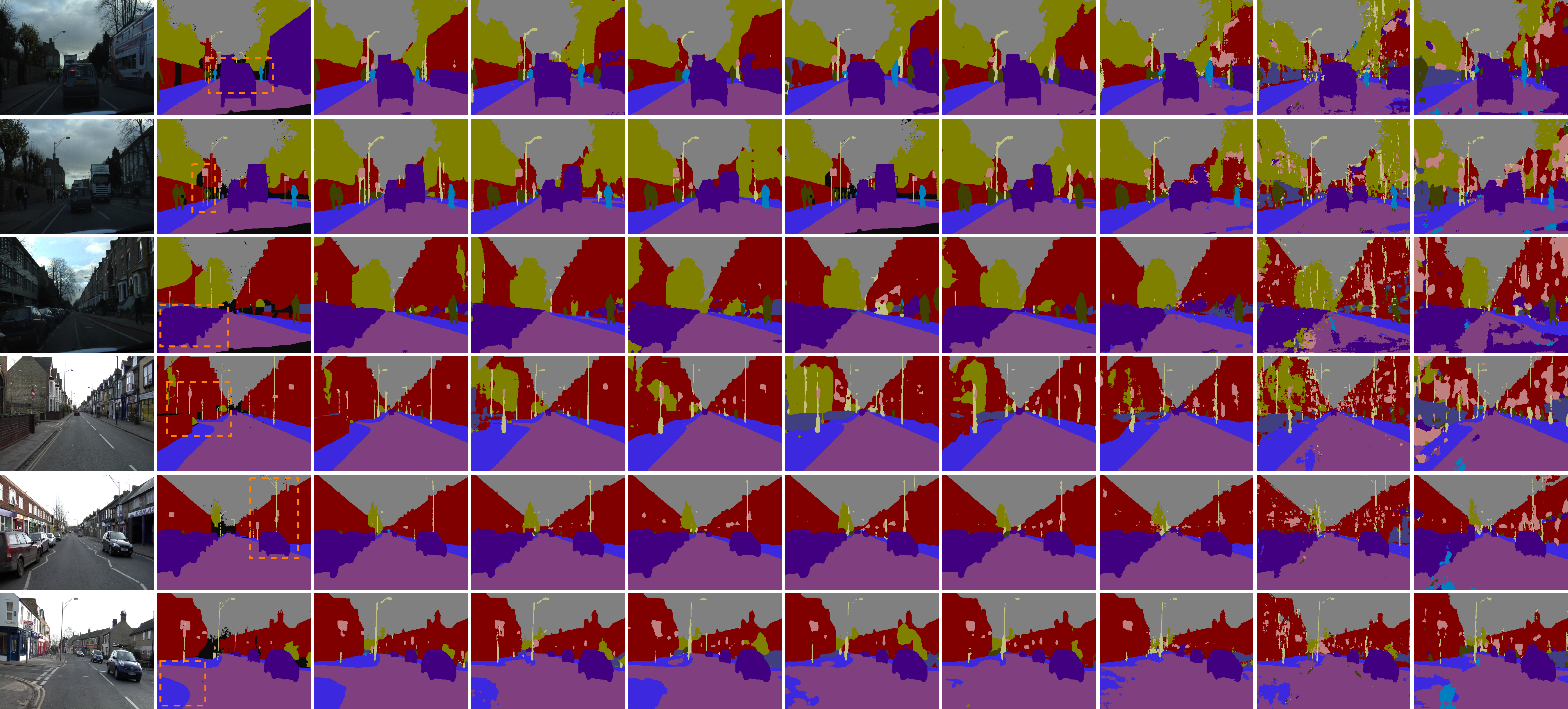}}
	\caption{Visual comparisons on the CamVid dataset. From left to right are original input images, ground truths, and segmentation results from our \textbf{LETNet}, FBSNet~\cite{gao2022fbsnet}, ICNet~\cite{zhao2018icnet}, DABNet~\cite{li2019dabnet}, BiseNet-v1~\cite{yu2018bisenet}, DFANet~\cite{li2019dfanet}, SegNet~\cite{badrinarayanan2017segnet}, and ENet~\cite{paszke2016enet}.}
	\label{Figure9}
\end{figure*}

\subsection{Ablation Study}
As depicted in Table~\ref{Table2}, some ablation experiments on the proposed modules are designed to prove the validity of these modules. \textbf{A. Long Connection}, \textbf{B. Feature Enhancement}, \textbf{C. Efficient Transformer}. Meanwhile, the baseline model is shown in Fig.\ref{Figure7}. It is worth noting that the baseline model does not introduce any proposed modules. The architecture of the baseline model is composed of an Encoder-Decoder composed of LDBs, which achieves 69.85\% mIoU on the validation set.

In group A, it is the effect of gradually adding L1, L2, L3, and it can be seen that after adding L1, there is a significant performance improvement of 0.58\%, proving that shallow information is greatly beneficial to the resolution of deep semantic information recovery. Meanwhile, when adding three long skip connections to the model, the performance increased by 1.16\% mIoU. This further verifies the importance of long connections for image segmentation.

In Group B, it is the effect of gradually adding L1, L2, and L3 to the network after joining Feature Enhancement. Comparing B1 and A1, it can be seen that adding FE only adds 101K parameters and 0.0032G FLOPs, but it can improve the performance of the model by 0.2\% mIoU, which is quite impressive.

In Group C, after introducing the Efficient Transformer, the model's performance has improved by 3.07\% mIoU, and the introduction of the pixel attention mechanism has also brought 0.5\% benefits to the model. Finally, C3* is the final version of the proposed LETNet, which improves the performance of the baseline model by 5.15\% mIoU.

All the above experiments fully demonstrate the necessity and effectiveness of the proposed modules and strategies.

\subsection{Comparisons with Advanced Models}

In this part, we compared recent years of representative semantic segmentation methods on the Cityscapes and CamVid datasets to demonstrate that our method strikes a better balance between segmentation accuracy and segmentation efficiency. 

\textbf{Evaluation on Cityscapes:} In Table~\ref{Table3},  we broadly divided the existing excellent methods into three categories: Large Size, Medium Size, and Small Size. Classification is based on parameters and calculations as the standard, the amount of parameters below 5M belongs to the Small Size category, and the calculation amount greater than 300G belongs to the Large Size category. It can be observed that the large-size models have obviously achieved outstanding segmentation effects, but their calculation complexity is high, the operation speed is slow, and they are not suitable for intelligent terminal hardware with high real-time requirements. 

In the medium size, the performance of HSBNet~\cite{li2021hierarchical} and LBN-AA~\cite{dong2020real} are slightly better than our LETNet. However, we should notice that the parameters of LBN-AA~\cite{dong2020real} and HSBNet~\cite{li2021hierarchical} are 6 times and 12 times larger than that of LETNet, respectively. 

In the small size category, our LETNet achieves the best results with fewer amount of parameters. This fully demonstrates that our LETNet can achieve a good balance between model size and performance. Indeed, BiseNet-v2~\cite{yu2021bisenet} is an outstanding model, which achieves similar mIoU results with slightly faster speed. However, we should not ignore that the number of parameters of LETNet is only 1/4 of BiseNet-v2. Meanwhile, Bisenet-v2 uses Xception as the backbone, which results in extra computational costs. In addition, we also list some of the methods detailed for each intersection classification over the union in Table~\ref{Table4}. Obviously, our LETNet achieves the best results in almost every class.

\textbf{Evaluation on CamVid:} In Table~\ref{Table5}, we provide a comparison of LETNet with other advanced methods on the CamVid dataset. According to the table, we can see that our LETNet still achieves the best result with only 0.95M parameters. This further verifies the effectiveness and excellence of the proposed LETNet.


\textbf{Visual Comparison}: In Figs.~\ref{Figure8} and~\ref{Figure9}, we also show the visual comparison of these methods on the Cityscapes and CamVid datasets, respectively. Obviously, our LEFTNet can get more accurate segmentation results. This is due to the well-design structure, and the ability of the Transformer can capture global correlation information, which helps to improve the accuracy of segmentation.

\section{Conclusion}
In this paper, we proposed a Lightweight Real-time Semantic Segmentation Network with Transformer and CNN. We combine the local feature extraction capabilities of CNNs with the long-range dependency modeling capabilities of Transformers. Specifically, an efficient Transformer is introduced in the middle of the model as a capsule network. Unlike the traditional Transformer, a more lightweight MHA is used, which can significantly reduce GPU memory consumption. Meanwhile, the Lightweight Dilated Bottleneck (LDB) module designed in CNN can learn more features under the premise of ensuring extreme simplicity and lightweight. Simultaneously, to make up for the shallow detail information lost by CNN in extracting deep semantic information, a U-shaped connection is used in the model. In connecting different levels, a Feature Enhancement (FM) module is also designed to improve the effective feature expression and suppress noise. Extensive experiment results show that our model makes an excellent balance between model size and performance.

\bibliographystyle{IEEEtran}
\bibliography{reference}

%

\begin{IEEEbiography}[{\includegraphics[width=1in,height=1.25in,clip,keepaspectratio]{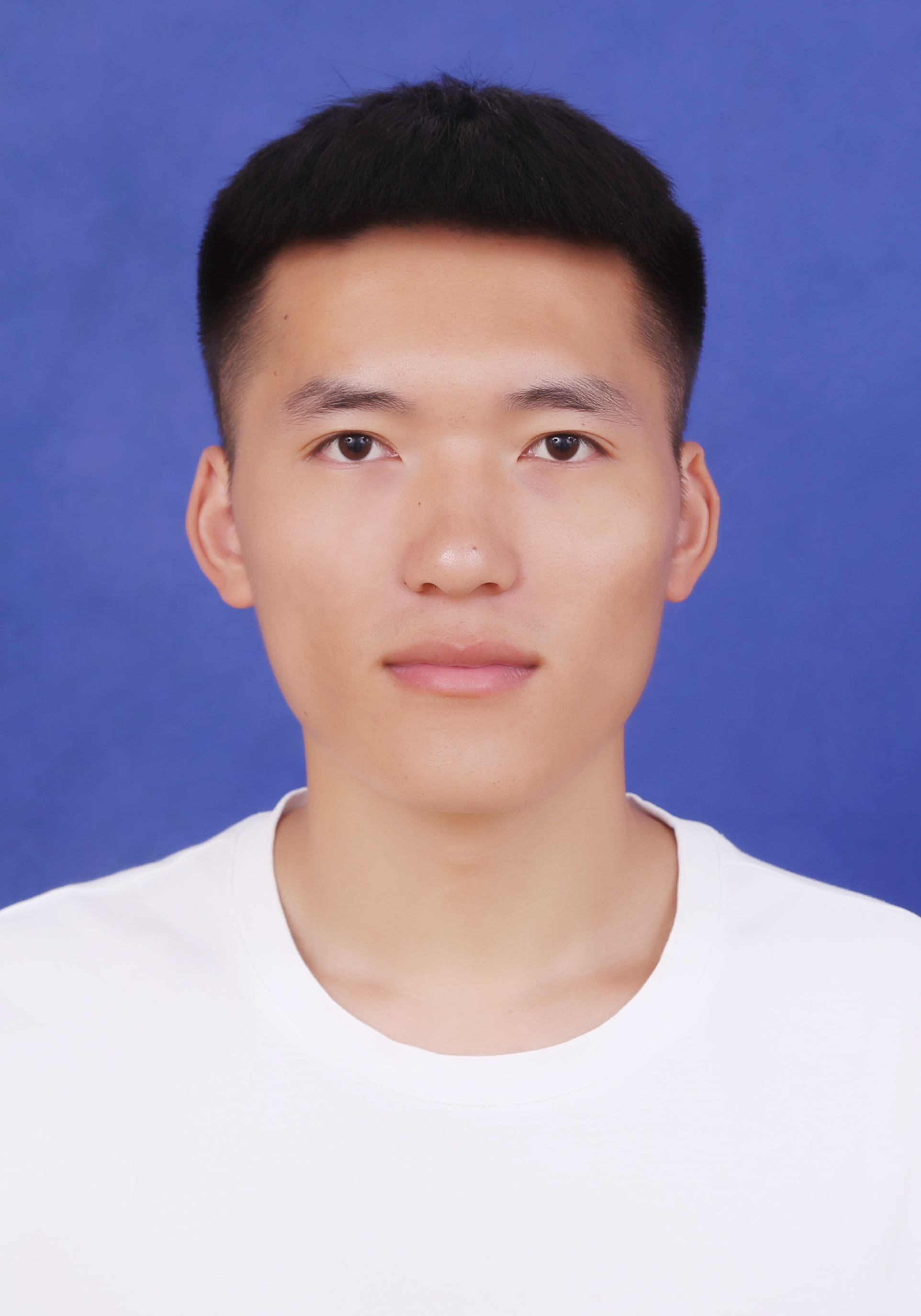}}]{Guoan xu}
received the B.S degrees in Measurement Control Technology and Instrumentation from Changshu Institute of Technology, Jiangsu, China, in 2019. He is currently pursuing the M.S. degree with the College of Automation \& College of Artificial Intelligence, Nanjing University of Posts and Telecommunications. His research interests include image semantic segmentation.
\end{IEEEbiography}

\begin{IEEEbiography}[{\includegraphics[width=1in,height=1.25in,clip,keepaspectratio]{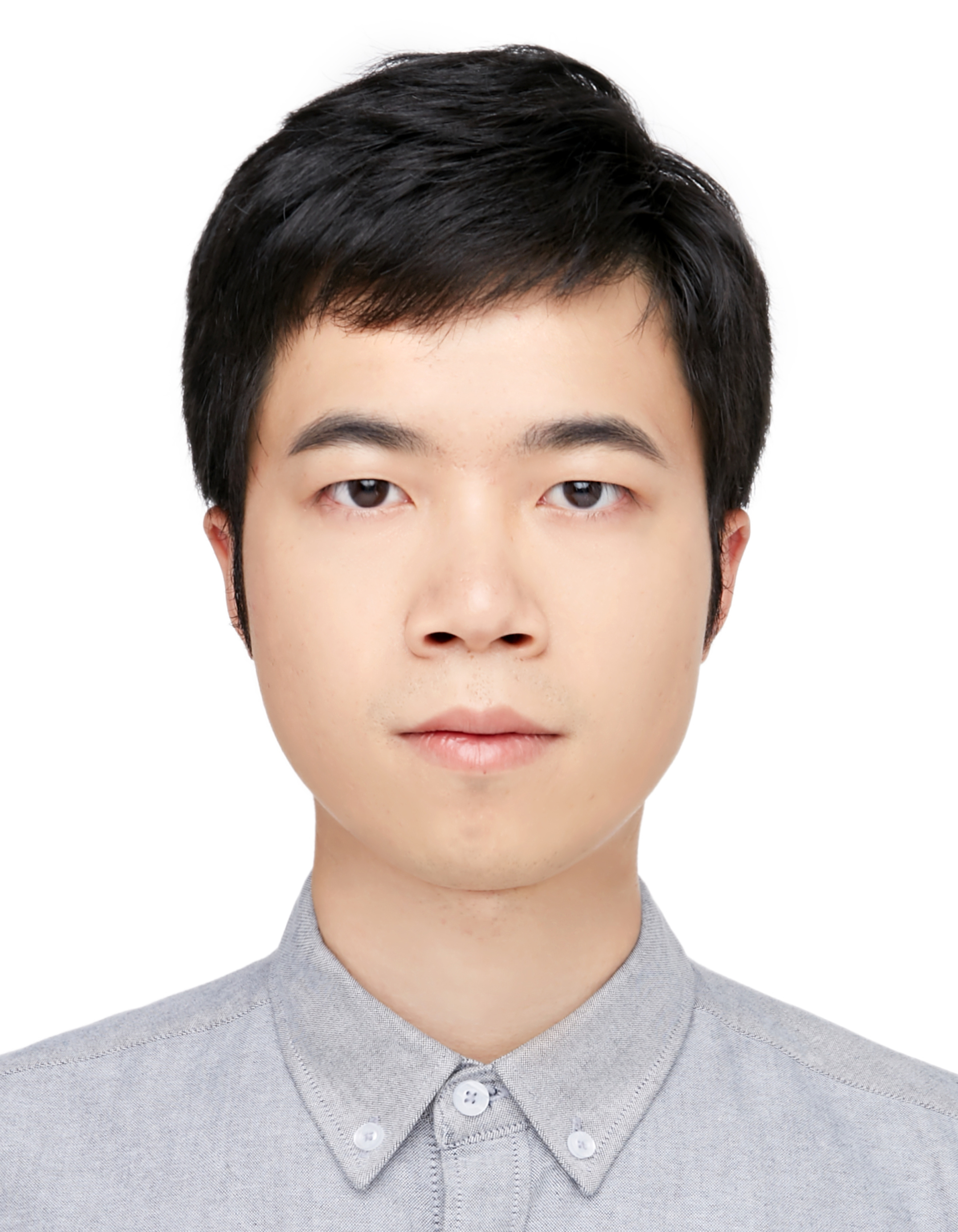}}]{Juncheng Li}
received the Ph.D. degree in Computer Science and Technology from East China Normal University, in 2021, and was a Postdoctoral Fellow at the Center for Mathematical Artificial Intelligence (CMAI), The Chinese University of Hong Kong. He is currently an Assistant Professor at the School of Communication \& Information Engineering, Shanghai University. His main research interests include image restoration, computer vision, and medical image processing. He has published more than 27 scientific papers in IEEE TIP, IEEE TNNLS, IEEE TMM, ICCV, ECCV, AAAI, and IJCAI.
\end{IEEEbiography}

\begin{IEEEbiography}[{\includegraphics[width=1in,height=1.25in,clip,keepaspectratio]{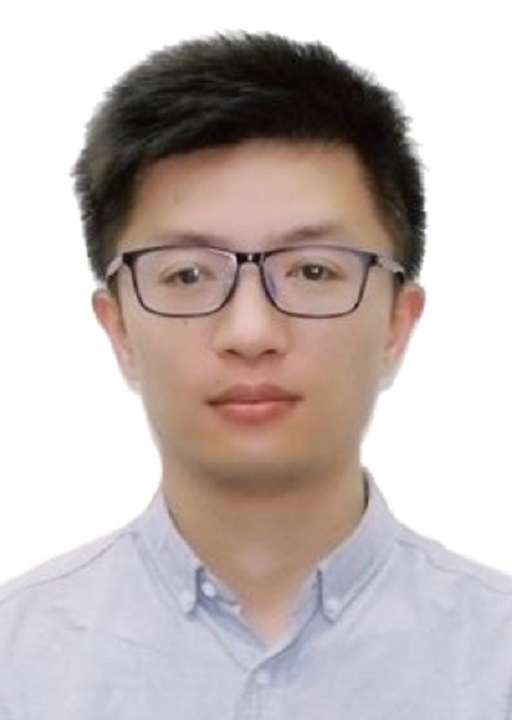}}]{Guangwei Gao}
(Senior Member, IEEE) received the Ph.D. degree in pattern recognition and intelligence systems from the Nanjing University of Science and Technology, Nanjing, in 2014. He was a Visiting Student of the Department of Computing, The Hong Kong Polytechnic University, in 2011 and 2013, respectively. He was also a Project Researcher with the National Institute of Informatics, Japan, in 2019. He is currently an Associate Professor in Nanjing University of Posts and Telecommunications. His research interests include pattern recognition and computer vision. He has published more than 60 scientific papers in IEEE TIP/TCSVT/TITS/TMM/TIFS, ACM TOIT/TOMM, AAAI, IJCAI, PR, etc. Personal website: \textit{https://guangweigao.github.io}.
\end{IEEEbiography}

\begin{IEEEbiography}[{\includegraphics[width=1in,height=1.25in,clip,keepaspectratio]{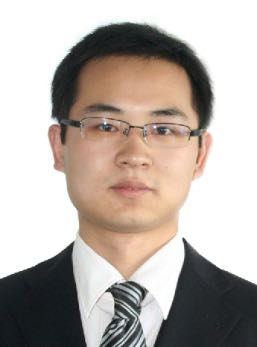}}]{Huimin Lu}
(Senior Member, IEEE) received the Ph.D. degree in electrical engineering from the Kyushu Institute of Technology in 2014. From 2013 to 2016, he was a JSPS Research Fellow (DC2, PD, and FPD) with the Kyushu Institute of Technology. He is currently an Assistant Professor with the Kyushu Institute of Technology and an Excellent Young Researcher of MEXT-Japan. His research interests include computer vision, robotics, artificial intelligence, and ocean observing.
\end{IEEEbiography}

\begin{IEEEbiography}[{\includegraphics[width=1in,height=1.25in,clip,keepaspectratio]{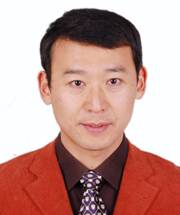}}]{Jian Yang}
(Member, IEEE) received the PhD degree from Nanjing University of Science and Technology (NUST), on the subject of pattern recognition and intelligence systems in 2002. In 2003, he was a postdoctoral researcher at the University of Zaragoza. From 2004 to 2006, he was a Postdoctoral Fellow at Biometrics Centre of Hong Kong Polytechnic University. From 2006 to 2007, he was a Postdoctoral Fellow at Department of Computer Science of New Jersey Institute of Technology. Now, he is a Chang-Jiang professor in the School of Computer Science and Engineering of NUST. His research interests include pattern recognition, computer vision and machine learning. Currently, he is/was an Associate Editor of Pattern Recognition Letters, IEEE Trans. Neural Networks and Learning Systems, and Neurocomputing. He is a Fellow of IAPR.
\end{IEEEbiography}

\begin{IEEEbiography}[{\includegraphics[width=1in,height=1.25in,clip,keepaspectratio]{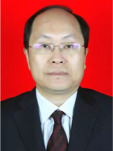}}]{Dong Yue}
(Fellow, IEEE) received the Ph.D. degree in engineering from the South China University of Technology, Guangzhou, China, in 1995. He is currently a Professor and Dean of the Institute of Advanced Technology and College of Automation \& AI, Nanjing University of Posts and Telecommunications. His current research interests include analysis and synthesis of networked control systems, multiagent systems, optimal control of power systems, and Internet of Things. Prof. Yue served as the Associate Editor for IEEE Industrial Electronics Magazine, IEEE Transactions on Industrial Informatics, IEEE Transactions on Systems, Man and Cybernetics: Systems, IEEE Transactions on Neural Networks and Learning Systems.
\end{IEEEbiography}





\end{document}